%% file: main.tex
\documentclass[10pt]{article}
\usepackage[preprint]{tmlr}

\usepackage{amsmath,amssymb}
\usepackage{booktabs}
\usepackage{array}
\newcolumntype{L}[1]{>{\raggedright\arraybackslash}p{#1}}
\usepackage{graphicx}
\usepackage{xcolor}
\usepackage{tikz}
\usetikzlibrary{positioning,arrows.meta,fit,calc}
\usepackage{microtype}
\usepackage{url}
\usepackage{enumitem}
\usepackage{float}
\usepackage{hyperref}

\definecolor{ours}{HTML}{2A78D6}
\definecolor{refc}{HTML}{EB6834}
\definecolor{inkii}{HTML}{52514E}
\hypersetup{colorlinks=true,bookmarksnumbered=true,linkcolor=ours,citecolor=ours,urlcolor=ours,
  pdftitle={Our Own Jev: Typed Decisions in One Forward Pass, from an In-House Judge to Open LoRA Models},pdfauthor={Yuchuan Weng, Chi Liu, Xin Chen, Srinivasan Manoharan}}

\newcommand{\WU}{W_{\!U}}

\title{Our Own Jev: Typed Decisions in One Forward Pass,\\from an In-House Judge to Open LoRA Models}

\author{\name Yuchuan Weng, Chi Liu, Xin Chen, Srinivasan Manoharan \\
\addr PayPal.AI}

\def\month{10}
\def\year{2026}
\def\openreview{}

\ifdefined\pdfminorversion\pdfminorversion=7\fi
\ifdefined\pdfinfoomitdate\pdfinfoomitdate=1\fi
\ifdefined\pdftrailerid\pdftrailerid{}\fi
\begin{document}

\maketitle

\begin{abstract}
A typed decision maps evidence, a question and a fixed set of options to a probability for every option,
computed in one forward pass, by reading a language model's logits or a small head over its hidden states,
instead of generating and parsing text.
The commercial system Jev popularised this System~1 interface, and open reproductions have followed. We
report two uses of the pattern. \emph{Part I} describes an in-house System~1 judge that triages about 148K
agentic coding session records for a training corpus. It replaces a frontier LLM judge that cost on the order of
\$1,000 per pass and whose stored labels matched a fresh two-judge consensus on only 26.9\% of the sessions
where the two judges agreed; a self-hosted frozen backbone with a rank-16 LoRA adapter and a pointer head
answers three typed questions in one forward pass, judges 1,000 sessions in about three minutes of one GPU,
about \$0.06 at a reference rate and roughly 1\% of the frontier judge's cost, and keeps every session inside
the organisation. \emph{Part II} asks whether a readout that adds no parameters can be competitive and whether training one
helps. Applying the public Eikos recipe and corpus of about 25K decisions to three further backbones
(1B, 2B and 35B), we read the frozen output layer of a LoRA-adapted decoder. On the 111 public hard items of
JevBench our single-seed 35B, developed with those items in view, scores 79.3\%, not significantly different from the commercial reference (73.0\%, unadjusted
$p=0.12$) and ahead of five open systems after Holm--Bonferroni correction; our 2B is ahead of all four kev
checkpoints on the leaderboard. A single-seed ablation finds no detectable benefit from training the 26 letter rows of the output layer, and a retrain
on a later corpus release moves the 2B's hard-tier accuracy by 11.7 points, which shows how little small evaluations
resolve.
\end{abstract}

\input{sections/01_introduction}
\input{sections/02_background}
\input{sections/partI_front}
\input{sections/03_method}
\input{sections/04_setup}
\input{sections/05_results}

\input{sections/06_readout}
\input{sections/10_provenance}
\input{sections/11_limitations}

\subsubsection*{Broader Impact Statement}
\input{sections/12_impact}

\subsubsection*{Acknowledgments}
We thank the author of Eikos for the open corpus, recipe and code, and the JevBench maintainer for the per-item records.

\nocite{brier1950verification,clopper1934use,cohen1960coefficient,hu2022lora,kahneman2011thinking,kev2026,li2024datacomp,penedo2024fineweb,zheng2023judging}
\bibliography{references}
\bibliographystyle{tmlr}

\appendix
\input{sections/partI_appendix}

\end{document}

%% file: sections/01_introduction.tex
\section{Introduction}
\label{sec:intro}

\begin{quote}
\emph{``System 1 operates automatically and quickly, with little or no effort and no sense of voluntary
control. System 2 allocates attention to the effortful mental activities that demand it.''}
\hfill --- \citet{kahneman2011thinking}
\end{quote}

Many operational decisions have a fixed set of allowed answers: which queue a ticket goes to, whether a
clause in a document is satisfied, whether a coding session is worth keeping as training data. Today they are
usually automated with System~2 machinery: prompt a chat model, ask it for JSON and parse the reply. That spends
hundreds of generated tokens on a choice among a handful of options, fails whenever the model writes prose instead
of an object, and returns a confidence value the model composes as text rather than computes.

A \emph{typed decision} interface is the System~1 alternative. Given evidence, a question and the
options, the model returns a probability for every option from a single forward pass, read from the model's outputs
(its logits, or a small head over its hidden states) rather than generated. The commercial system Jev
\citep{typesafe2026jev} popularised this interface, and JevBench \citep{jevbench} benchmarks systems that implement
it on accuracy, calibration, speed and cost. Because the product is closed, an open-source race followed, with
reproductions built on encoders, frozen base models, pointer heads and new readouts (Section~\ref{sec:background});
a fifth project, Eikos \citep{eikos2026}, releases a corpus, a training recipe, an evaluation harness and models that
read letter logits without adding parameters.

This paper reports two efforts with the same pattern in two settings.

\textbf{Part I: an in-house System~1 judge (\S\ref{sec:deploy-task}--\ref{sec:deploy-system}).} We
collect agentic coding sessions, about 148,000 session records and growing, as training data for an in-house
coding model, and each must be triaged by three typed questions before it enters the corpus. A frontier
LLM judge cost on the order of \$1,000 per pass, needed about 73 hours of judging time, and its stored labels
had drifted to the point that, where two fresh judges agreed, their label matched the stored one on only
26.9\% of sessions; the data also could not leave the organisation. A self-hosted judge built on the kev
architecture (a frozen 4B or 27B backbone, a rank-16 LoRA adapter and a pointer head) answers all three
questions in one forward pass, judges 1,000 sessions in about three minutes of one GPU, about \$0.06 at a
reference rate and roughly a hundredth of the frontier judge's cost, and reaches 83--85\% (coding),
91--92.5\% (in-domain) and 68--69.5\% (eight-way domain) agreement with frontier-judge labels on a fixed,
class-balanced held-out set of 200 sessions (1.5 to 3.8 points lower on the 144 of them that share no state
with the training records).

\textbf{Part II: open typed-decision models (\S\ref{sec:method}--\ref{sec:provenance}).} Part~II asks whether a readout that adds no parameters can be competitive, and whether training one helps. We read the \emph{existing} output layer of a decoder language
model at the single position where the answer letter would be generated, restrict it to the letters in play, and
train only a LoRA adapter \citep{hu2022lora} with soft cross-entropy. The prompt format (SemIf's), the training
recipe, the evaluation battery and the corpus of about 25K public decisions are those of the open Eikos project, which
pairs a letter-logit readout with task training; ours are the adapters for three further backbones (a 1B, a 2B and a
35B mixture of experts), the paired comparisons on JevBench's 111 public hard items and a single-seed test of
whether training the readout helps. The 35B reaches 79.3\%: after Holm--Bonferroni correction it is significantly
ahead of five open systems, but not significantly different from the commercial reference (73.0\%, unadjusted
$p=0.12$). Training the readout buys nothing we can measure, and a
retrain of the 2B on a later corpus release moved its hard-tier accuracy from 61.3\% to 73.0\% although the two runs
differ in more than the seed (Sections~\ref{sec:readout} and~\ref{sec:provenance}).

All JevBench numbers in Sections~\ref{sec:setup}--\ref{sec:readout} are on the public half (version 1.4.2.2); the
sealed half is scored only by its operator. Section~\ref{sec:provenance} discloses provenance and which builds were
developed with the public items in view.

%% file: sections/02_background.tex
\section{Background and related work}
\label{sec:background}

\paragraph{Typed decisions and JevBench.}
An item consists of a state $s$ (free text or structured evidence), a question $q$ of one of three types
(boolean, categorical choice, or ordinal score), and options $O=(o_1,\dots,o_K)$; a system returns
$p\in\Delta^{K-1}$. JevBench \citep{jevbench}, in version 1.4.2.2, publishes 231 items in three tiers (easy 48,
standard 72, hard 111); the rest, including a judge tier, are held by the operator and scored only on submission.

\paragraph{Open reproductions.}
Table~\ref{tab:landscape} summarises the open reproductions whose code we read; they differ mainly in where the
decision is computed. Laya \citep{laya2026} is a ModernBERT encoder \citep{warner2024modernbert}. SemIf \citep{semif2026} reads option-letter logits from an unmodified Qwen3.5-4B, with a
structured prompt that Eikos later adopted. kev \citep{kev2026} adds a pointer head and a block-causal option-isolation
mask that lets one pass answer several questions (Part~I builds on kev). imajev \citep{imajev2026} adds a 256-code
readout, copied from the output layer and then trained on about a million decisions. Eikos \citep{eikos2026}, the
source of Part~II's recipe, data and evaluation battery, reads the same letter logits with no added parameters from
LoRA-adapted Qwen backbones trained on its own corpus of about 25K decisions. The other systems we compare, from
the JevBench leaderboard (``the board''), are reflex \citep{reflex2026} (a Qwen model read by direct logits),
Winnow \citep{winnow2026} (a LoRA-tuned Gemma), djev \citep{djev2026} (an inference method on a diffusion language
model), jqv \citep{jqv2026} (a zero-shot Qwen3-32B) and OpenJev Verdict \citep{verdict2026} (a 151M encoder). None of
Laya, SemIf, kev and imajev combines task training with a readout that adds no parameters; Eikos does (as do other
board entries such as reflex-4B), and we build on that idea rather than claim it. Part~II supplies evidence on other
backbones, paired comparisons with a selection of the field, and a test of whether the readout needs training at all.

\begin{table}[t]
\centering
\small
\caption{Open typed-decision reproductions, as described by each project's published code and
documentation at the time of writing, and Part~II's.}
\label{tab:landscape}
\begin{tabular}{@{}L{1.25cm}L{3.1cm}L{3.1cm}L{3.0cm}L{3.6cm}@{}}
\toprule
& Backbone & Readout & Training & Stated or observed limit \\
\midrule
Laya & ModernBERT-large (421M) & softmax over option markers & own recipe & 512-token budget \\
SemIf & Qwen3.5-4B, unmodified & option-letter logits & none & score questions excluded \\
kev & Qwen decoder & new pointer head & LoRA + head from scratch & mask not applied on hybrid backbones \\
imajev & Qwen3.5 2B/4B/9B & 256-code layer copied from the output layer & LoRA, 5 stages, $\sim$1M decisions & raw hard-tier calibration error 0.113 \\
Eikos & Qwen3.5-4B, Qwen3.8-27B & option-letter logits, frozen output layer & LoRA, soft targets, $\sim$25K decisions & self-reported hard tier 72.1 (4B), 82.9 (27B) \\
\textbf{Ours} (Part~II) & MiniCPM5-1B/2B, Qwen3.6-35B-A3B & frozen output layer; \textbf{no new parameters} & Eikos recipe and corpus & 26 letters per pass \\
\bottomrule
\end{tabular}
\end{table}

\paragraph{Related techniques.}
Scoring multiple-choice answers by the probability of an option symbol is standard in LLM evaluation
\citep{hendrycks2021mmlu,robinson2023leveraging} and can be well calibrated \citep{kadavath2022language}; models are
sensitive to option order and label bias \citep{zheng2024selectors,pezeshkpour2024order,zhao2021calibrate}, which
motivates our per-example reshuffling.
We measure calibration with expected calibration error \citep{naeini2015obtaining,guo2017calibration}, consider
temperature scaling and report error under selective prediction \citep{elyaniv2010foundations,geifman2017selective}.
Soft targets follow knowledge distillation \citep{hinton2015distilling}, the rationale loss follows
\citet{hsieh2023distilling}, and we adapt with LoRA through PEFT \citep{mangrulkar2022peft}. Our analysis draws on work on
significance testing and power in NLP \citep{card2020little,dror2018hitchhiker,bergkirkpatrick2012empirical}, error
bars and small test sets \citep{miller2024adding,bowyer2025position} and seed and numerical variance
\citep{dodge2020finetuning,bouthillier2021accounting,madaan2024quantifying,summers2021nondeterminism,zhuang2022randomness},
and uses McNemar's exact test \citep{mcnemar1947note,dietterich1998approximate}, the paired bootstrap
\citep{efron1993bootstrap} and the Holm--Bonferroni correction \citep{holm1979simple}. Our contamination check is an
$n$-gram overlap test in the tradition of \citet{brown2020language}, with the caveats of \citet{sainz2023nlp}. The judge
of Part~I replaces an LLM-as-a-judge \citep{zheng2023judging}.

%% file: sections/03_method.tex
\section{Part II. Open typed-decision models: readout, recipe and data}
\label{sec:method}

The judge of Part~I reads its answer through kev's trained pointer head. Part~II asks whether a readout that
adds no parameters at all can be competitive, and whether training one helps. The recipe, the corpus and the
evaluation battery are those of the open Eikos project \citep{eikos2026}, whose released models already read letter
logits this way; Part~II adds adapters for three further backbones, paired comparisons with the commercial system
and a selection of the open field on JevBench, and a single-seed test of whether training the readout helps
(\S\ref{sec:readout}). It does not compare a pointer head with the letter readout under matched conditions.
The two parts share an interface and a lesson, not an experiment: Part~I shows what the pattern buys in a deployment,
Part~II tests the readout on a public benchmark, and in both the evaluation sets (200 sessions, 111 items) are too
small to resolve the differences that matter.

\subsection{A readout with no new parameters}

\begin{figure}[b]
\centering
\begin{tikzpicture}[
  font=\small, node distance=4mm and 6mm,
  box/.style={draw=inkii, rounded corners=2pt, align=center, inner sep=4pt, line width=0.5pt},
  frozen/.style={box, fill=black!5},
  train/.style={box, draw=ours, fill=ours!10, line width=0.8pt},
  arr/.style={-{Stealth[length=2mm]}, line width=0.6pt, draw=inkii}]
\node[box, text width=3.3cm] (inp) {\textbf{Prompt}\\[1pt]\footnotesize evidence $s$ \; criterion $q$\\ options \texttt{A}\,$o_1$ \; \texttt{B}\,$o_2$ \; \texttt{C}\,$o_3$\\ \textit{answer position $t$}};
\node[frozen, right=of inp, text width=2.9cm] (bb) {\textbf{Decoder backbone}\\[1pt]\footnotesize all weights frozen};
\node[train, below=2.5mm of bb, text width=3.4cm] (lora) {\footnotesize LoRA $\Delta W{=}BA$, rank 64,\\ on every linear layer};
\node[frozen, right=of bb, text width=2.6cm] (wu) {\textbf{Output layer $\WU$}\\[1pt]\footnotesize the model's own, frozen;\\ no new parameters};
\node[box, right=of wu, text width=2.5cm] (out) {\footnotesize logits of \texttt{A},\texttt{B},\texttt{C}\\ $\rightarrow$ softmax\\ $\rightarrow$ $p(o_1),p(o_2),p(o_3)$};
\draw[arr] (inp) -- (bb);
\draw[arr] (bb) -- node[above,font=\footnotesize]{$h_t$} (wu);
\draw[arr] (wu) -- (out);
\draw[arr, draw=ours] (lora) -- (bb);
\node[below=3mm of out, text width=3.3cm, align=center, font=\footnotesize, text=inkii] (loss) {training: soft cross-entropy\\ to target distribution $y$\\ $+\,0.3\times$ rationale LM loss};
\draw[arr, dashed] (out) -- (loss);
\end{tikzpicture}
\caption{The model. One forward pass; the decision is read from the backbone's own output layer at a
single position, restricted to the letters in play. Only the LoRA matrices (blue) are trained; the
rationale loss is used in training only.}
\label{fig:arch}
\end{figure}
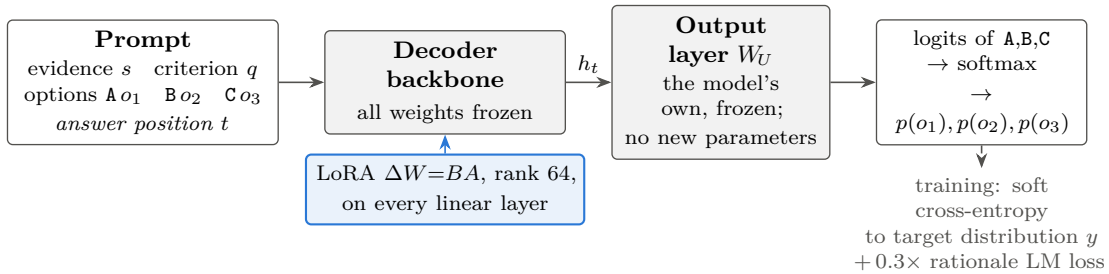

We render the state, the question and the options into a prompt in which option $k$ is introduced
by a letter token $\ell_k\in\{\texttt{A},\texttt{B},\dots\}$, and end the prompt exactly where the
answer letter would be generated. The prompt is SemIf's structured format \citep{semif2026} as adopted by Eikos:
a JSON object with the evidence, the criterion and a list of options, each with a letter and a description, under a
system instruction to answer with the uppercase letter of one option. Let $h_t(\theta)$ be the final hidden state at
that position for a backbone with frozen weights and LoRA parameters $\theta$, and let $\WU$ be the backbone's
own, frozen, unembedding matrix. The decision is
\begin{equation}
\label{eq:readout}
p_\theta(o_k\mid s,q,O)\;=\;\frac{\exp\big(z_{\ell_k}\big)}{\sum_{j=1}^{K}\exp\big(z_{\ell_j}\big)},
\qquad z=\WU\,h_t(\theta).
\end{equation}
We call this the \emph{letter readout} (Figure~\ref{fig:arch}). Only $\theta$ is trained. Boolean, choice and ordinal score questions share
it. With single-letter labels, the configuration we use, one pass handles up to 26 options; larger sets are
answered by a tournament over blocks of at most 26 whose winners are asked again.

Equation~\ref{eq:readout} is the standard way of reading a class probability from a pretrained output layer, as in the
verbalizers of prompt-based fine-tuning \citep{schick2021exploiting,gao2021making} and monoT5's ``true''/``false''
logits \citep{nogueira2020document}, and it keeps the backbone's calibration over those letters. kev learns a pointer
head from scratch and imajev trains a new $256\times d$ layer initialised from the rows of $\WU$; SemIf and Eikos add no
parameters. Section~\ref{sec:readout} tests training the 26 letter rows.

\subsection{Recipe and data}
\label{sec:recipe}

The recipe is Eikos's and we tuned none of it: LoRA adapters of rank 64 on every linear projection, one epoch,
learning rate $10^{-4}$, effective batch 32, soft cross-entropy against a full target distribution over the options, an
auxiliary loss on a short written rationale (weight 0.3, never used at inference \citep{hsieh2023distilling}) and
option order re-permuted per example. The 1B and 2B follow Eikos's 4B settings (its first run, without weight
averaging or auxiliary terms) and add a symmetric KL divergence (weight 0.5) between the option distributions of an item
and of its English or Brazilian-Portuguese view; the 35B follows its 27B settings (no views, a 14,000-token context).
Apart from the readout we ablated none of these choices. The data is the public \texttt{eikos-decisions} corpus \citep{eikos2026} (CC BY 4.0), as cloned on 2026-09-24:
programmatically generated items with computed answers, teacher distributions, derivatives of FinEntity
\citep{tang2023finentity}, TAT-QA and GSM8K training data, long-context dossiers and the views. Our 1B and 2B used
a snapshot of 25,130 rows (a task family, a topic and a language held out whole, 1,191 rows) plus 1,619 view rows;
our 35B used 24,447 rows. A subset was retracted by the corpus author after our snapshot was taken
(Section~\ref{sec:provenance}). Inference is one forward pass and a softmax over at most 26 logits at temperature 1;
several questions about one state share a prefix, as in Eikos's serving code
\citep{kwon2023pagedattention,zheng2024sglang}.

%% file: sections/04_setup.tex
\section{Experimental setup}
\label{sec:setup}

\paragraph{Models and benchmark.} We adapt MiniCPM5-1B, MiniCPM5-2B \citep{hu2024minicpm} and Qwen3.6-35B-A3B
\citep{yang2025qwen3} with Eikos's settings by size (Section~\ref{sec:recipe}); each is
trained once, with a single seed, and the 2B is released as a merged standalone model \citep{pjev2026}. We score the
231 public JevBench items (commit \texttt{fd54ea7}, v1.4.2.2; later versions are not rerun) and recompute competitors from its published per-item records, so comparisons are paired (52
systems have such records; we use 15, selected by us; imajev-4b publishes aggregates only). To measure what 111
items cannot, we also use the seven-suite battery of the Eikos repository (7,371 items: the JevBench public items,
DecisionBench, Eikos's template-generated trade and rulebook suites, which make up 31\%, a general battery and two
finance suites) and its policy of deciding at confidence $\geq0.9$.

\paragraph{Statistics.} Paired comparisons use the exact McNemar test, and 95\% intervals are exact Clopper--Pearson
intervals (mapped to the difference for paired comparisons), except the ECE interval, a 20,000-sample bootstrap.
Because we report many paired tests, we apply the Holm--Bonferroni correction \citep{holm1979simple} within the
comparisons we report for each model (11 for the 35B, 15 for the 2B and 3 for the 1B, whose comparison with the
commercial reference is descriptive; our selection, not preregistered), call a difference significant only if its
adjusted $p$ is below 0.05, and call the rest not significant, which on 111 items is not evidence of equivalence. A
111-item tier has 80\% power only for differences of about ten points between two runs of one recipe (twelve to
twenty between unrelated systems, which disagree on more items).

%% file: sections/05_results.tex
\section{Results on JevBench}
\label{sec:results}

\subsection{Accuracy and paired comparisons}

Table~\ref{tab:tiers} gives hard-tier accuracy for 15 of the 52 systems whose outcomes the benchmark publishes (our
selection; imajev-4b is added from its published aggregates) and the paired differences of our 35B and 2B from each.
Of the 37 systems we did not compare, 11 score higher than our 2B (up to 76.6\%) and none higher than our 35B.
The 35B is significantly ahead of five open systems (Laya, kev 8B, reflex-4B, SemIf and jqv, the last two untrained)
and behind the two frontier LLMs by 17.1 points. Against the commercial reference it is $+6.3$ points (unadjusted
$p=0.119$), so on 111 items we can say neither that it is better nor that the two are equivalent; so does djev, Winnow-12B and reflex-27B. The 35B's five ``ahead'' rows do not depend on the choice of Holm family (the largest raw $p$ is 0.0003). The 2B is significantly ahead of Laya and of all four kev checkpoints, significantly behind reflex-27B
and the frontier LLMs, and not significantly different from SemIf, jqv and reflex-4B (60--61\% on the hard tier); the 1B is
significantly ahead of kev 4B and Laya (a family of three, excluding the reference) and trails the commercial
reference by 18.9 points. Three of the 2B's verdicts sit at the edge: kev 8B, kev 0.6B and reflex-27B would not survive a correction over all 52 systems (adjusted
$p=0.148$, 0.154 and 0.182), whereas ``ahead of every kev checkpoint'', read as a conjunction, needs only that each
raw $p$ is below 0.05 (the largest is 0.0055). Part~II's first question, whether a readout with no new parameters can
be competitive, therefore has only a weak answer: the 35B is significantly ahead of five open systems and not shown to
be worse than the commercial reference.

Accuracy rises with size across our builds (54.1, 61.3 and 79.3) but not across kev's checkpoints (29.7, 43.2, 36.9
and 45.0 at 0.5B, 0.6B, 4B and 8B; the 8B is two items ahead of the 0.6B, $p=0.87$), which is the evidence behind
Part~I's remark that kev scales poorly. Both patterns are suggestive only: our builds use different backbones and
one seed each, the 1B--2B gap is below what a 111-item tier can resolve (Section~\ref{sec:readout}), and kev's
checkpoints are of a previous generation, with a Qwen2.5 base at 0.5B (the Qwen3.5-based kev of Part~I is not on the
board); their 0.6B--4B inversion is seven items (exact McNemar $p=0.28$). SemIf's 61.3 from an unmodified 4B is what
letter logits give without task training, which our 2B matches at about 60\% of the parameters; since the 35B's lead
over the untrained reflex-27B is not significant, these data do not show how much of its score comes from training.
Two board systems run the same Qwen3.6-35B-A3B backbone (SimpleJev, 65.8, direct logits, and openjev-sglang, 73.0; the
board describes no training for either); our 79.3 is 13.5 and 6.3 points above them, with no 35B per-item scores to
pair. Eikos's own models report 72.1 (4B) and 82.9 (27B) in its README, without per-item records, 11 and 4 points above our
2B and 35B, so we do not claim that the recipe transfers without loss.

\begin{table}[t]
\centering
\small
\caption{Hard-tier accuracy (111 public items, v1.4.2.2) and paired differences of our 35B and 2B from each system
(points; Holm-adjusted $p$ in parentheses, within each model's reported comparisons: 11 for the 35B, 15 for the 2B;
$^*$ adjusted $p<0.05$; -- not in that model's family). ``Training'' is what the board or the project's documentation
says is trained. 2B rows are recomputed from per-item scores; 35B rows come from the original evaluation (merged rows share one recorded
difference).}
\label{tab:tiers}
\begin{tabular}{@{}llrll@{}}
\toprule
System & Training & Hard & 35B vs.\ system & 2B vs.\ system \\
\midrule
DeepSeek Flash; GPT-5.6 Luna & generative & 96.4 & $-17.1$ ($<$.002)$^*$ & $-35.1$ ($<$.0001)$^*$ \\
\textbf{Ours 35B-A3B} & LoRA & \textbf{79.3} & & \\
reflex-27B & none & 75.7 & $+3.6$ (.48) & $-14.4$ (.049)$^*$ \\
Winnow-12B & LoRA & 73.0 & $+6.3$ (.42) & $-11.7$ (.18) \\
Jev 1.13.0 & closed & 73.0 & $+6.3$ (.36) & $-11.7$ (.088) \\
imajev-4b & LoRA, readout & 72.1 & -- & -- \\
djev & none & 67.6 & $+11.7$ (.096) & $-6.3$ (1.0) \\
jqv; SemIf 4B & none & 61.3 & $+18.0$ (.002)$^*$ & $0.0$ (1.0) \\
\textbf{Ours 2B} & LoRA & \textbf{61.3} & & \\
reflex-4B & LoRA & 60.4 & $+18.9$ ($<$.002)$^*$ & $+0.9$ (1.0) \\
\textbf{Ours 1B} & LoRA & \textbf{54.1} & & \\
kev 8B & LoRA, head & 45.0 & $+34.2$ ($<$.002)$^*$ & $+16.2$ (.046)$^*$ \\
kev 0.6B & LoRA, head & 43.2 & -- & $+18.0$ (.046)$^*$ \\
OpenJev Verdict 1.4 & encoder, trained & 36.9 & -- & $+24.3$ (.001)$^*$ \\
kev 4B & LoRA, head & 36.9 & -- & $+24.3$ (.0002)$^*$ \\
Laya & encoder, trained & 35.1 & $+44.1$ ($<$.002)$^*$ & $+26.1$ (.003)$^*$ \\
kev 0.5B & LoRA, head & 29.7 & -- & $+31.5$ ($<$.0001)$^*$ \\
\bottomrule
\end{tabular}
\end{table}

\subsection{Calibration and selective decisions}

For a decision system the operative question is often how wrong it is when it is confident enough to act without
review. Table~\ref{tab:selective} reports, on the 7,371-item battery, the share of decisions made at confidence
$\geq0.9$ and the error among them: the 35B commits on 42.3\% of items and is wrong on 1.9\% of those. Two cautions
apply. First, up to 31\% of the battery is template siblings of the training data, where the models are far more accurate:
for the 2B the error among confident decisions is 4.2\% on the full battery but 7.7\% (95\% interval 6.0--9.7) on the
four external suites we could score per item. Second, the commercial system's 5.9\% is Eikos's figure on six suites
that exclude the JevBench public items and we did not rerun Jev, so the two are not comparable. On the hard tier
our expected calibration error (ECE) is 0.084 (35B), 0.090 (1B) and 0.120 (2B), against imajev-4b's published 0.082 after
its fitted temperature, but ECE on 111 items is noisy (a bootstrap 95\% interval for the 2B's is $[0.079, 0.208]$). A fitted temperature does not help our larger models: it comes out below one (0.80
for the 35B, 0.95 for the 2B) and worsens hard-tier ECE on held-out halves of the public items. No held-out family, topic or language
collapses toward chance at any size (Table~\ref{tab:selective}, bottom).

\begin{table}[t]
\centering
\small
\caption{Top: selective decisions and ECE on the 7,371-item battery, and for the 2B on the four external suites
we could score per item (4,403 items); Jev's figure is Eikos's, on six other suites, and not comparable. Bottom: categorical held-out sets, accuracy (\%) / ECE.}
\label{tab:selective}
\begin{tabular}{@{}lrrr@{}}
\toprule
& Decided at $\geq$0.9 & Error among those & ECE \\
\midrule
Ours 35B & 42.3\% & 1.9\% & 0.028 \\
Ours 2B & 27.0\% & 4.2\% & 0.021 \\
Ours 1B & 6.9\% & 4.3\% & 0.060 \\
Ours 2B, external suites & 19.1\% & 7.7\% & -- \\
Jev 1.13.0 & -- & 5.9\% & -- \\
\bottomrule
\end{tabular}
\quad
\begin{tabular}{@{}lrrr@{}}
\toprule
Held out & 35B & 2B & 1B \\
\midrule
Family: tradeoff & 93.6 / .074 & 85.9 / .051 & 71.8 / .064 \\
Topic: healthcare adm. & 96.1 / .050 & 90.5 / .056 & 81.6 / .072 \\
Language: Spanish & 93.5 / .045 & 89.0 / .041 & 77.7 / .057 \\
\bottomrule
\end{tabular}
\end{table}

%% file: sections/06_readout.tex
\section{Does the readout need to be trained?}
\label{sec:readout}

imajev-4b trains its readout rows and ships a frozen-rows fallback without
comparing them. We ran that comparison as a single-variable ablation at two sizes, which also shows how much two
single-seed runs can differ.

\paragraph{Design.} For MiniCPM5-1B and 2B we trained two arms with identical data, schedule and seed. In
\emph{trained}, the letter rows of $\WU$ are copied into a separate fp32 linear layer and unfrozen; in \emph{frozen},
the same copied layer stays frozen. The arms differ only in whether those rows receive gradients (the readout gradient also enters the global
clipping norm); the trained rows moved 5.4\% (2B) and 4.1\% (1B) in relative norm. This tests the mechanism, not imajev's shipped configuration (256 rows, 655K parameters at
4B, about a million decisions, five stages): ours updates 26 rows (53K parameters at 2B) on about 23K decisions.

\begin{table}[t]
\centering
\small
\caption{Trained vs.\ frozen readout rows, one seed per arm (hard tier: 111 items; battery: 7,371 paired items).
Control rows: frozen arm vs.\ released model of the same size (same readout rows; fp32 linear vs.\ bf16 gather; different
trainer file); 1B interval not recomputed.}
\label{tab:readout}
\begin{tabular}{@{}llrrrrr@{}}
\toprule
& & \multicolumn{2}{c}{Accuracy (\%)} & & & \\
\cmidrule(lr){3-4}
Size & Set & trained & frozen & $\Delta$ (pp) & 95\% CI & $p$ \\
\midrule
2B & hard tier & 63.1 & 67.6 & $-4.5$ & $[-11.0, +3.6]$ & 0.332 \\
1B & hard tier & 48.6 & 54.1 & $-5.4$ & $[-11.2, +2.5]$ & 0.210 \\
2B & battery & 70.6 & 70.2 & $+0.41$ & $[-0.43, +1.24]$ & 0.348 \\
1B & battery & 59.2 & 58.8 & $+0.41$ & $[-0.40, +1.21]$ & 0.330 \\
\midrule
2B & battery, control & \multicolumn{2}{c}{frozen vs.\ released} & $-1.11$ & $[-1.88, -0.34]$ & 0.0046 \\
1B & battery, control & \multicolumn{2}{c}{frozen vs.\ released} & $+0.62$ & -- & 0.059 \\
\bottomrule
\end{tabular}
\end{table}

\paragraph{Result.} Table~\ref{tab:readout} shows no detectable effect. On the hard tier the trained readout is
nominally worse at both sizes, but neither difference is significant, and that tier's minimum detectable effect is
about ten points. On the 7,371-item battery the differences are $+0.41$ points at both sizes, with intervals whose upper ends are $+1.24$
and $+1.21$ points, which reflect item sampling only, not the variation between training runs (below). With one seed per arm, training the readout buys
no accuracy we can measure; that is a statement about benefits of about two points or more, and about this 26-row
version of the idea.

\paragraph{Run-to-run floor.} A frozen arm and the released model of the same size have identical readout rows but
compute the logits through an fp32 matrix product rather than a bf16 gather (identical in exact arithmetic, not
numerically), come from different trainer files and share a seed. After 708 steps they differ by $-1.11$ points on the
battery at $p=0.0046$ (2B) and $+0.62$ at $p=0.059$ (1B), and by $+6.3$ and $0.0$ points on the hard tier. McNemar's
test conditions on items but not on the variance between training trajectories, which here is confounded with the
treatment. Two pairs cannot estimate run-to-run spread, but near-identical runs can differ by about a
point on the battery and several points on the hard tier, so we treat pooled-battery differences below about 1.5
points as unresolved regardless of $p$.

\paragraph{The hard-tier floor.} A 2B retrained with Eikos's 27B settings on the corpus release that applies the
retraction of Section~\ref{sec:provenance} (seed 7) scores 73.0\% on the hard tier against 61.3\% for the released
2B: $+11.7$ points (95\% interval $[+2.6, +17.6]$, exact McNemar $p=0.011$). The runs differ in more than the seed
(corpus release, code version of the imported Eikos modules, batch shape, context limit, and a loss with no view rows
or view term), so the difference cannot be attributed to the seed, to the retraction or to any one other change. The
2B-class runs we have (released 61.3, trained readout 63.1, frozen readout 67.6, an exploratory ordinal-loss arm 66.7,
the retrain 73.0) span 61.3 to 73.0. Two pilot arms on that release added 2,400 generated drill rows (designed after
inspecting public-item failures) or 1,739 CC BY sentiment rows and score 65.8 and 62.2, 7.2 and 10.8 below the
retrain, so data changes alone moved the tier by 7 to 11 points. We read hard-tier differences below about ten points
between 2B-class systems as unresolved.

\paragraph{Relation to Part I.} Part~I (Section 4.4) found that a pointer head trained from scratch on other questions
answered the untrained eight-way question at about chance (14\% for the 27B; 4B heads initialised from kev's checkpoint
scored 52--59\%). A readout with no new parameters has no such head, but we did not test the eight-way question with it; the held-out family, topic and language of Table~\ref{tab:selective} are the closest
evidence that the letter readout transfers to question types unseen in training.

%% file: sections/10_provenance.tex
\section{Provenance, contamination and selection}
\label{sec:provenance}

\paragraph{Training data and a retraction.} All training data of the Part~II models is the public
\texttt{eikos-decisions} corpus (CC BY 4.0), apart from 1.7--2.4K public or programmatically generated rows added
in two of the further runs of Section~\ref{sec:readout}; no proprietary or internal data and no outputs of the
commercial system were used. On 2026-09-27 the corpus author removed 1,994 FinEntity-derived rows,
because those documents are test items of another benchmark suite (not the DecisionBench of our battery), and asked that they be excluded from future training. Our snapshots were taken on
2026-09-24, so every Part~II model, including the released 2B, was trained on them. We report no result on that
suite, and an exact match of all 968 documents against five of our evaluation suites found no overlap. We retrained the 2B only without them. The retrain loses 23.0 points on the
300-item financial-sentiment task of the battery's finance suite (71.7\% to 48.7\%), a task closely related to what the
retracted rows teach, with no significant change on the other two finance tasks (pooled $-2.5$ points, 95\% interval
$[-6.6,+1.6]$), DecisionBench or the general battery, and gains 11.7 points on the JevBench hard tier
(Section~\ref{sec:readout}). The finance numbers in this paper therefore very probably depend on the retracted
rows; the 1B and the 35B have not been retrained.

\paragraph{Contamination.} We applied the corpus author's published check (word 8-grams over the state and the
question instructions, containment in both directions, flagged above 0.15). Against the 231 public items, no row of
the 1B/2B snapshot or of the 35B snapshot is flagged; the maximum containment is 0.0398 and positive controls fire. The check is lexical: it cannot
detect paraphrase, translation or regeneration from a shared template, five public items are too short for it, and the teacher model that produced the soft
labels may itself have seen the benchmark, a channel we cannot measure. The corpus also deliberately covers the
benchmark's task taxonomy.

\paragraph{Selection.} The released 2B was trained on 2026-09-24, before the benchmark was first downloaded to our
machines; its development set is a hash split of our own corpus, it was trained once on a fixed schedule with no
checkpoint selection, and it ships at temperature 1. The same cannot be said of the 1B or the 35B (both were trained or selected after public-item scores of earlier builds
had been seen), of the drill arm of Section~\ref{sec:readout} (its rows were designed after inspecting public-item
failures) or of further builds that we trained and do not report. We report the exact-recipe 35B, chosen when both 35B builds' public scores were known (79.3 versus
80.2 on hard), launched on 2026-09-27, the day of the retraction, from the snapshot of 2026-09-24. Their public-item
numbers are therefore optimistic in the usual way. Nor are the recipe and the corpus independent of the public items:
in its own JevBench submission the corpus author states that the 231 public items were one of the project's
development evaluation sets, used to choose its final checkpoint, and that they were never trained on. Our
hyperparameters are the upstream ones, so the exposure is indirect. The comparison systems were
likewise selected by us from the 52 with published per-item records.

\paragraph{Reproducibility.} Part~II uses only public inputs (the Eikos corpus and code as cloned on 2026-09-24, the base
models, JevBench's public items and per-item records). Our released 2B is a merged standalone model
\citep{pjev2026}; the 1B and 35B adapters and their per-item scores are not released. Part~I's inputs are internal session records and cannot be
released (Section~\ref{sec:deploy-task}). Appendix F is the only appendix (Part~I's lettering and Tables 14--16).

%% file: sections/11_limitations.tex
\section{Limitations}
\label{sec:limitations}

\begin{itemize}[leftmargin=1.4em,itemsep=1pt,topsep=2pt]
\item \textbf{Part I evidence.} The in-house judge uses kev's pointer head rather than the readout of
  Part~II. It is evaluated against the labels of a single frontier judge, which also confirmed the bank
  strata, on 200 class-balanced held-out sessions, 56 of which overlap the training records (the conclusions
  hold without them). The domain question covers eight of fourteen legacy categories, with no ``other''
  option. Numbers come from the deployment's own harness; self-hosted costs are GPU time priced at a reference
  rate, and full-pool figures are projections from measured latency.
\item \textbf{Borrowed recipe, data and evaluation.} The training recipe, the corpus, the prompt format and
  the seven-suite evaluation battery with its confidence-0.9 decision policy are Eikos's. Besides the readout we
  ablated only an exploratory ordinal-loss arm and two data-addition runs (Section~\ref{sec:readout}), and our adapters inherit the corpus's coverage, its teacher and its licence mix.
\item \textbf{No matched comparison and no untrained baselines of our own.} We did not train a pointer head or
  an encoder on the same backbone and data, so the comparison with kev is across checkpoints that differ in base
  model, data and training, and we did not score our own backbones untrained, so what the adapters add is not
  measured (two board systems run the 35B's backbone, Section~\ref{sec:results}). Nor did we train a cheap non-LLM
  classifier on Part~I's rendered state, the natural cost competitor to the typed judge, so Part~I does not show that
  the typed interface is needed for its savings. The ablation tests a 26-row version of imajev's readout, not its shipped configuration.
\item \textbf{No sealed measurement.} Every JevBench number above is on the public half of v1.4.2.2, and systems score
  far lower on the sealed half than on the public items; we report no sealed result.
\item \textbf{One seed and selection.} The observed difference between near-identical runs is about one point on a 7,371-item battery
  (1.1 and 0.6 in two pairs; we treat differences below 1.5 as unresolved)
  and, from one retrain that differs in more than the seed, about ten points on the hard tier; effects below those
  sizes are unresolved, including the readout ablation. The 1B
  and the 35B were developed with the public items visible, the upstream recipe was itself exposed to them indirectly
  (\S\ref{sec:provenance}), and we chose the comparison systems and multiplicity families from 52 systems with
  per-item records.
\item \textbf{Retracted training rows.} All Part~II models were trained on since-retracted rows; only the 2B was
  retrained without them, in a run that differs in more than the data.
\item \textbf{Evaluation suites.} The trade and rulebook suites are template siblings of training data (up to 31\% of the
  battery), so battery-wide confident-error rates are lower than on external suites. The commercial reference appears
  only through JevBench's per-item records and, for confident-decision error, Eikos's published figure. The raw $p$-values of the 35B and 1B comparisons come from the original
  evaluation, whose per-item files are not released; the Holm adjustments were applied here. The 35B needs about 80\,GB of memory.
\end{itemize}

\section{Conclusion and next steps}

The typed-decision pattern holds up in both settings we studied. In-house (Part~I), a self-hosted
System~1 judge turns a projected \$1,000, 73-hour LLM-judge pass over 148K coding session records into about
seven GPU-hours, without any session leaving the organisation, at accuracy that 200 held-out sessions cannot
distinguish between a 4B and a 27B backbone. In the open (Part~II), reading a language model's own output
layer in one forward pass, with a LoRA adapter trained on about 25K public decisions under the Eikos recipe and no
new parameters, gives a 35B that is significantly ahead of five open systems (two without task training) on
the hard tier and not significantly different from the commercial reference, and a 2B that is ahead of all four kev
checkpoints on the board; training 26 letter rows on top of the output layer bought nothing we could measure at
one seed. Small evaluation sets bound what either part can claim: a 111-item tier has 80\% power only for differences
of about ten points between two runs of one recipe and a retrain on a later corpus release (with other changes) moved
our 2B by 11.7 points on it.

Three steps follow. First, a sealed evaluation of our 2B, together with a multi-seed replication of the readout ablation and a
retraining of every model without the retracted rows. Second, the 2B rather than the 35B as the candidate deployable
open model, once retrained without the retracted rows: its single-question latency on one H100 is about a quarter of
the 35B's (20.5 against 83.2 ms). Third, pointing the open recipe at our own
decisions: other bounded, high-volume decisions in the organisation (for example triage, routing and
categorisation) share the evidence, question and options shape of Part~I, and there we will compare the zero-parameter letter readout against the pointer
head directly.

%% file: sections/12_impact.tex
Typed-decision models are built to automate high-volume decisions, some of which affect people: eligibility,
prioritisation and routing of requests. Calibrated probabilities and the low error rate we measure
among confident decisions make it tempting to let such systems act without review. Those error rates
are measured on benchmark items, not on any deployment distribution, and Section~\ref{sec:readout}
shows how little a small public tier can resolve. Deployments should
re-measure calibration on their own traffic, keep humans in the loop for consequential decisions,
and treat the confidence threshold as a policy choice. On the positive side, self-hosting keeps
sensitive inputs within the organisation that holds them and makes model versions auditable.

%% file: sections/partI_appendix.tex
\PIfirst{8}{733.5}{3.64}{
\PIhead{72}{-569.25}{1}{F Part I: inputs, training configuration and overlap check}{appendix.F}
\PIlabel{72}{-569.25}{app:partI}{F}{appendix.F}}
\PIcont{9}{164.5}{0}{
\PIlink{180.4}{-154.53}{11.95}{10.79}{table.14}}
\PInext{10}{317.6}{0}{
\PIdest{114.41}{-170.36}{table.14}}
\PInext{11}{408}{0}{
\PIlink{221.91}{-338.33}{11.86}{10.89}{table.15}}
\PInext{12}{514.2}{0}{
\PIdest{116.71}{-413.94}{table.15}}
\PInext{13}{550.3}{0}{
}
\PInext{14}{601.8}{0}{
\PIlink{366.1}{-579.91}{14.98}{10.79}{subsection.4.5}
\PIlink{529.09}{-579.91}{12.15}{10.79}{table.16}
\PIlink{133.08}{-591.86}{6.97}{10.79}{table.2}}
\PInext{15}{716}{0}{
\PIdest{119.78}{-607.69}{table.16}}